\documentclass{article}

\usepackage{arxiv}
\usepackage{amsmath}
\usepackage{amsthm}
\usepackage{xcolor}
\usepackage{enumerate}

\usepackage[utf8]{inputenc} 
\usepackage[T1]{fontenc}    
\usepackage{hyperref}       
\usepackage{url}            
\usepackage{booktabs}       
\usepackage{amsfonts}       
\usepackage{nicefrac}       
\usepackage{microtype}      
\usepackage{lipsum}
\usepackage{float}
\usepackage{caption} 
\usepackage{subcaption}
\usepackage{mathtools}
\usepackage{float}
\usepackage{booktabs,siunitx}
\usepackage{makecell}
\usepackage{multirow}

\newcommand{\norm}[1]{\left\lVert#1\right\rVert}
\newtheorem{theorem}{Theorem}

\title{Orthogonal-Pad\'e Activation Functions: Trainable Activation functions for smooth and faster convergence in deep networks}

\author{
  Koushik Biswas
  \\
   \And
 Shilpak Banerjee \\
  \And
  Ashish Kumar Pandey \\
}

\begin{document}
\maketitle

\begin{abstract}
We have proposed orthogonal-Pad\'e activation functions, which are trainable activation functions and show that they have faster learning capability and improves the accuracy in standard deep learning datasets and models. Based on our experiments, we have found two best candidates out of six orthogonal-Pad\'e activations, which we call safe Hermite-Pade (HP) activation functions, namely HP-1 and HP-2. When compared to ReLU, HP-1 and HP-2 has an increment in top-1 accuracy by 5.06\% and 4.63\% respectively in PreActResNet-34, by  3.02\% and 2.75\% respectively in MobileNet V2 model on CIFAR100 dataset while on CIFAR10 dataset top-1 accuracy increases by 2.02\% and 1.78\% respectively in PreActResNet-34, by 2.24\% and 2.06\% respectively in LeNet, by 2.15\% and 2.03\% respectively in Efficientnet B0.
\end{abstract}


\section{Introduction}
Deep networks are constructed with multiple hidden layers and neurons. Non-linearity is introduced in the network via activation function in each neuron. ReLU \cite{relu} is proposed by Nair
and Hinton and is the favourite activation in the deep learning community due to its simplicity. Though ReLU has a drawback called dying ReLU, and in this case, up to 50\% neurons can be dead due to vanishing gradient problem, i.e. there are numerous neurons which has no impact on the network performance. To overcome this problem, later Leaky Relu \cite{lrelu}, Parametric ReLU \cite{prelu}, ELU \cite{elu}, Softplus \cite{softplus} was proposed, and they have improved the network performance though it's still an open problem for researchers to find the best activation function. Recently Swish \cite{swish} was found by a group of researchers from Google brain, and they used automated searching technique. Swish has shown some improvement in accuracy over ReLU. GELU \cite{gelu}, Mish \cite{mish}, TanhSoft \cite{tanhsoft}, EIS \cite{eis} are few other candidates proposed recently which can replace ReLU and Swish.

In the recent past, there is an increasing interest in trainable activation function. Trainable activation functions have learnable hyperparameter(s), which are updated during training via backpropagation algorithm\cite{backp}. In this paper, we have proposed Orthogonal-Pad\'e activation functions. Orthogonal-Pad\'e functions can approximate most of the continuous functions.

\section{Pad\'e activation Unit (PAU) and Orthogonal-PAU}
Consider a close interval $[a,b]$ of the real line. Let $\mathcal{P}_n(x)$ be the space of all polynomials in $x$ of degree smaller or equal to $n$. For a non-negative, continuous function $w(x)$ on $[a,b]$, define an inner product on $\mathcal{P}_n(x)$ as
\begin{align}\label{eq:inner}
    <P,Q>_w=\int_a^b w(x)P(x)Q(x)~dx.
\end{align}
A finite set of polynomials $\{P_1(x),P_2(x),\cdots,P_k(x)\}$ is said to be orthogonal if 
\begin{align}
    <P_i,P_j>_w=0  \quad \text{ if } i\neq j.
\end{align}
A basis for $\mathcal{P}_n(x)$ is a set of $n$ polynomials whose span is whole of $\mathcal{P}_n(x)$. An orthogonal basis is a basis that is also an orthogonal set.  

A standard basis for $\mathcal{P}_n(x)$ is $\{1, x, x^2, \cdots, x^n\}$. 
But the standard basis is not orthogonal with respect to the inner product defined in \eqref{eq:inner}. In many applications, working with an orthogonal basis simplifies expressions and reduce calculations. There are several well known orthogonal basis for the space of Polynomials. Table~\ref{tab1} enlists some of these polynomial bases. Note that some of them are given by recurrence relations and others by direct expressions. 

\begin{table}[H]
    \centering
    \begin{tabular}{|c|c|c|}
        \hline
         {\bf Polynomial} &  {\bf Recurrence Relation/Expression}\\
         \hline
         Chebyshev polynomial of the first kind (CP-1) & $r_0(x)=1,r_1(x)=x,r_{n+1}(x)=2xr_n(x)-r_{n-1}(x)$ \\
         \hline
         Chebyshev polynomial of the Second kind (CP-2) & $r_0(x)=1,r_1(x)=2x,r_{n+1}(x)=2xr_n(x)-r_{n-1}(x)$\\
         
         \hline
         Laguerre polynomials (LAU) & $r_0(x)=1,r_1(x)=1-x,r_{n+1}(x)=\frac{(2n+1-x)r_n(x)-nr_{n-1}(x)}{n+1}$\\
         \hline
         Legendre polynomials (LEG) & $r_n(x) = \sum_{k=0}^{[n/2]}(-1)^k\frac{(2n-2k)!}{2^nk!(n-2k)!(n-k)}x^{n-2k}$\\
         \hline
         Probabilist's Hermite polynomials (HP-1) & $\mathit {r}_{n}(x)=(-1)^{n}e^{\frac {x^{2}}{2}}{\frac {d^{n}}{dx^{n}}}e^{-{\frac {x^{2}}{2}}}$\\
         \hline
         
         Physicist's Hermite polynomials (HP-2) &  $r_{n}(x)=(-1)^{n}e^{x^{2}}{\frac {d^{n}}{dx^{n}}}e^{-x^{2}}$\\
         \hline
    \end{tabular}
    \vspace{0.2cm}
    \caption{Some well-known Orthogonal Polynomial Bases.}
    \label{tab1}
\end{table}

\subsection{Pad\'e activation Unit (PAU)}
The Pad\'e approximation of $f(x)$ by a rational function $F_1(x)$ is defined as 
\begin{align}\label{eq2}
    F_1(x) = \frac{P(x)}{Q(x)} = \frac{\sum_{i=0}^{k}a_ix^i}{1+\sum_{j=1}^{l}b_jx^j} = \frac{a_0+a_1x+a_2x^2+\cdots+a_kx^k}{1+b_1x+b_2x^2+\cdots+b_lx^l}
\end{align}
where $P(x)$ and $Q(x)$ are polynomials of degree $k$ and  $l$ respectively and they have no common factor. PAU \cite{pau} is a learnable activation function of the form given in \eqref{eq2} where the polynomial coefficients $a_i, b_j, 0\leq i\leq k, 1\leq j\leq l$ are learnable parameters and updated during back-propagation.
To remove the pole of $F_1(x)$ coming from zeros of $Q(x)$, authors in \cite{pau} proposed safe PAU. Safe PAU is defined as
\begin{align}\label{eq3}
    F_2(x) = \frac{P(x)}{Q(x)} = \frac{\sum_{i=0}^{k}a_ix^i}{1+|\sum_{j=1}^{l}b_jx^j|} = \frac{a_0+a_1x+a_2x^2+\cdots+a_kx^k}{1+|b_1x+b_2x^2+\cdots+b_lx^l|}
\end{align}
Introducing the absolute value in the denominator ensures that the denominator will not vanish. In fact, one can take absolute value inside the sum and define
\begin{align}\label{eq4}
    F_3(x) := \frac{P(x)}{Q(x)} = \frac{\sum_{i=0}^{k}a_ix^i}{1+\sum_{j=1}^{l}|b_j||x^j|} = \frac{a_0+a_1x+a_2x^2+\cdots+a_kx^k}{1+|b_1||x|+|b_2||x^2|+\cdots+|b_l||x^l|}
\end{align}
We will show that in many tasks activation functions defined by $F_3$ provide better results than safe PAU defined in $F_2$.

\subsection{Orthogonal-Pad\'e activation Unit (OPAU)}The orthogonal-Pad\'e approximation of $g(x)$ by a rational function $G(x)$ is defined as 
\begin{align}\label{eq5}
    G(x) = \frac{P(x)}{Q(x)} = \frac{\sum_{i=0}^{k}c_if_i(x)}{1+\sum_{j=1}^{l}djf_j(x)} = \frac{c_0+c_1f_1(x)+c_2f_2(x)+\cdots+c_kf_k(x)}{1+d_1f_(x)+d_2f_2(x)+\cdots+d_lf_l(x)}
\end{align}
where $f_t(x)$ belongs to a set of orthogonal polynomials (see \cite{ortho}). As in the case of PAU, the learnable activation function, OPAU, is defined by \eqref{eq5} where $c_i, d_j, 0\leq i\leq k, 1\leq j\leq l$ are learnable parameters. The parameters are initialized by taking approximation of the form \eqref{eq5} of a well-known activation function like ReLU, Leaky ReLU etc., see \cite{pau}. To remove poles of $G(x)$, we propose safe OPAU as follows (abusing notation)
\begin{align}\label{eq6}
    G(x) = \frac{P(x)}{Q(x)} = \frac{\sum_{i=0}^{k}c_if_i(x)}{1+\sum_{j=1}^{l}|dj||f_j(x)|} = \frac{c_0+c_1f_1(x)+c_2f_2(x)+\cdots+c_kf_k(x)}{1+|d_1||f_1(x)|+|d_2||f_2(x)|+\cdots+|d_l||f_l(x)|}
\end{align}
We have considered six orthogonal polynomial bases - Chebyshev (two types), Hermite (two types), Laguerre, and Legendre polynomial bases for this work. Details about these polynomial bases are in Table~\ref{tab1}. 

\subsection{Learning activation parameters via back-propagation} Weights and Biases in neural network models are updated via backpropagation algorithm and gradient decent. The same method is adopted to update the activation parameters. We have implemented the forward pass in both Pytorch \cite{pytorch} \& Tensorflow-Keras \cite{keras} API and automatic differentiation will update the parameters. Alternatively, CUDA \cite{cuda} based implementation (see \cite{pau}, \cite{lrelu}) can be used and the gradients of equations (\ref{eq5}) for the input $x$ and the parameters $c_i$'s and $d_j$'s can be computed as follows:
\begin{align}
    \frac{\partial G}{\partial x} = \frac{1}{Q(x)}\frac{\partial P(x)}{\partial x} - \frac{P(x)}{Q(x)^2}\frac{\partial Q(x)}{\partial x},\ \ \
    \frac{\partial G}{\partial c_i} = \frac{f_i(x)}{Q(x)}, \ \ \
    \frac{\partial G}{\partial d_j} = -\operatorname{sgn}(d_j)|f_j(x)|\frac{P(x)}{Q(x)^2}.
\end{align}
\section{Networks with orthogonal-Pad\'e activations and function approximation}
Orthogonal-Pad\'e networks are similar to Pad\'e networks \cite{pau} in which a network with PAU or safe PAU is replaced with an OPAU or safe OPAU. In this article, we have considered safe OPAUs as an activation function with different orthogonal bases as given in Table~\ref{tab1}. We have initialized the learnable parameters (Polynomial coefficients) using the approximation of Leaky ReLU by the functional form given in \eqref{eq6}, see Table~\ref{tab10} for initializing parameter values. The network parameters have been optimized via the backpropagation method \cite{backp}. We have kept a similar design for all networks as PAU in \cite{pau}, for example, weight sharing and learning activation parameter per layer \cite{weightshare}. From equation (5), we have a total $(k+l)$ extra parameters per layer. So if there are L layers in a network, there will be extra $L\times (k+l)$ numbers of learnable parameters in the network. To train a network, we have adopted Leaky ReLU initialization ($\alpha = 0.01$) (see appendix for details) instead of the random initialization method, and results are reported in the experiments section.

A major advantage of using an orthogonal basis is that the polynomial coefficients can be found uniquely much faster in running time compared to a standard basis. Also, Widely used activation functions in most cases are zero centered. We impose some conditions on Pad\'e and Orthogonal-Pad\'e approximation to make the known function approximation (we have considered Leaky ReLU initialization) zero centered and check whether there is any advantage (one definite advantage is the number of parameters reduces in each layer) on model performance. To make Pad\'e zero centered, we replace $a_0=0$ in equation (4) and calculate the rest other parameters. For safe OPAU, several cases arrive, and we explore all possible cases. For example, if we choose HP-1 as a basis, the safe OPAU function approximation can be zero centered if the constant term in the numerator is zero. So, we have from equation (6) and Table~\ref{tab1},  $c_0-c_2+3c_4=0$. The following cases can be derived:

Case-1: $c_0=c_2=c_4=0$.

Case-2: one of $c_0$ or $c_2$ or $c_4$ is equal to zero. Example, if $c_0=0$, then $c_2=3c_4$ etc.

Case-3: $c_0=c_2-3c_4$ or $c_2=c_0+3c_4$ or $c_4=\frac{1}{3}(c_2-c_0)$.

In all the above cases for PAU and HP-1, the rational approximation to Leaky ReLU has been explored and tested on several models on CIFAR10 \cite{cifar10} and CIFAR100 \cite{cifar10} datasets. We find that in most cases, model performance in top-1 accuracy reduces by 0.2\%-0.6\%.

Also, note that the class of neural networks with safe OPAU activation functions is dense in $C(K)$, where $K$ is a compact subset of $\mathbb{R}^n$ and $C(K)$ is the space of all continuous functions over $K$.\\
The proof follows from the following propositions (see \cite{pau}). 

\textbf{Proposition (Theorem 1.1 in Kidger and Lyons, 2019 \cite{universal}) :-} Let $\rho: \mathbb{R}\rightarrow \mathbb{R}$ be any continuous function. Let $N_n^{\rho}$ represent the class of neural networks with activation function $\rho$, with $n$ neurons in the input layer, one neuron in the output layer, and one hidden layer with an arbitrary number of neurons. Let $K \subseteq \mathbb{R}^n$ be compact. Then $N_n^{\rho}$ is dense in $C(K)$ if and only if $\rho$ is non-polynomial.

\textbf{Proposition 2. (From Theorem 3.2 in (Kidger and Lyons, 2019)):-} Let $\rho: \mathbb{R}\rightarrow \mathbb{R}$ be any continuous function which is continuously differentiable at at least one point, with nonzero derivative at that
point. Let $K \subseteq \mathbb{R}^n$ be compact. Then $NN_{n,m,n+m+2}^\rho$ is dense in $C(K; \mathbb{R}^m)$.

\section{Experimental results with Orthogonal-Pad\'e Activation}
We have first initialized the trainable parameters(polynomial coefficients) of safe OPAU activation functions by rational function approximation of Leaky ReLU ($\alpha=0.01$) activation and then updated the parameters via backpropagation algorithm via (7). The coefficients are given in appendix A in table~\ref{tab10}. In the next subsections, we have given details of our experimental setup, experimental results on different deep learning problems like image classification, Object detection, Machine Translation in some widely used standard datasets. We have considered ReLU \cite{relu}, Leaky ReLU \cite{lrelu}, ELU \cite{elu}, Softplus \cite{elu}, and Swish \cite{swish} as our baseline activation functions to compare performance on different networks with safe OPAU activations.
\subsection{MNIST}The MNIST \cite{mnist} is a popular computer vision database contains handwritten digits from 0 to 9. The database has a total of 60k training and 10k testing $28\times 28$ grey-scale images. We have used a custom 8-layer homogeneous convolutional neural network (CNN) architecture with $3\times 3$ kernels on CNN layers and pooling layers with $2\times 2$ kernels. Channel depths of size 128 (twice), 64  (thrice), 32 (twice), a dense layer of size 128, Max-pooling layer(thrice) are used with batch-normalization\cite{batch}, and dropout\cite{dropout}. We have not used the Data augmentation method. The results are reported in Table~\ref{tab2}.

\begin{table}[H]
\begin{center}
\begin{tabular}{ |c|c|c| }
 \hline
 Activation Function &  \makecell{5-fold mean accuracy (\%) \\ on MNIST test data}  \\
 \hline
 ReLU  &  99.14  \\ 
 \hline
 Leaky ReLU($\alpha$ = 0.01) & 99.22  \\
 \hline
 ELU  & 99.15\\
 \hline
 Softplus & 99.01\\
 \hline
 Swish  & 99.23 \\
 \hline
 PAU & 99.24\\
 \hline
 CP-1 & 99.22\\
 \hline
 CP-2 & 99.27\\
 \hline
 LAU & 99.25\\
 \hline
 LEG & 99.20\\
 \hline
 HP-1 & \textbf{99.47}\\
 \hline
 HP-2& \textbf{99.40}\\
 \hline

 \end{tabular}
 \vspace{0.2cm}
\caption{Comparison between different baseline activations and  safe OPAU activations on MNIST dataset.} 
\label{tab2}
\end{center}
\end{table}
\subsubsection{\textbf{Fashion MNIST}}
Fashion-MNIST \cite{fashion} is an image database consisting of $28\times 28$ grey-scale images. The database consists of a total of ten classes of fashion items with 60k training images and 10k testing images. We have not used any Data augmentation method. The same CNN network which is used for the MNIST dataset is also used for this database and, the results are reported in table ~\ref{tab3}.

\begin{table}[H]
\begin{center}
\begin{tabular}{ |c|c|c| }
 \hline
 Activation Function &  \makecell{5-fold mean accuracy (\%) \\ on Fashion MNIST test data}  \\
 \hline
 ReLU  &  92.87  \\ 
 \hline
 Leaky ReLU($\alpha$ = 0.01) & 92.91 \\
 \hline
 ELU  & 92.97\\
 \hline
 Softplus & 92.78\\
 \hline
 Swish  & 92.99\\
 \hline
 PAU & 93.05\\
 \hline
 CP-1 & 93.02\\
 \hline
 CP-2 & 93.01\\
 \hline
 LAU & 93.15\\
 \hline
 LEG & 93.09\\
 \hline
 HP-1 & \textbf{93.39}\\
 \hline
 HP-2 & \textbf{93.31}\\
 \hline

 \end{tabular}
 \vspace{0.2cm}
\caption{Comparison between different baseline activations and  safe OPAU activations on Fashion MNIST dataset.} 
\label{tab3}
\end{center}
\end{table}
\subsubsection{\textbf{The Street View House Numbers (SVHN) Database}}
SVHN \cite{SVHN} consists of $32\times 32$ RGB images of real-world house numbers of Google's street view images. There are total 73257 training images and  26032 testing images, and the images are spread over 10 different classes. The same CNN network which is used for the MNIST dataset is also used for this database and, the results are reported in table~\ref{tab4}. We have used the data augmentation method.
\begin{table}[H]
\begin{center}
\begin{tabular}{ |c|c|c| }
 \hline
 Activation Function &  \makecell{5-fold mean accuracy (\%) \\ on SVHN test data}  \\
 \hline
 ReLU  &  95.17  \\ 
 \hline
 Leaky ReLU($\alpha$ = 0.01) & 95.22  \\
 \hline
 ELU  & 95.19\\
 \hline
 Softplus & 95.01\\
 \hline
 Swish  & 95.27 \\
 \hline
 PAU & 95.29\\
 \hline
 CP-1 & 95.25\\
 \hline
 CP-2 & 95.23\\
 \hline
 LAU & 95.30\\
 \hline
 LEG & 95.21\\
 \hline
 HP-1 & \textbf{95.43}\\
 \hline
 HP-2 &  \textbf{95.42}\\
 \hline

 \end{tabular}
 \vspace{0.2cm}
\caption{Comparison between different baseline activations and  safe OPAU activations on SVHN dataset.} 
\label{tab4}
\end{center}
\end{table}
\subsubsection{\textbf{CIFAR}}
\begin{figure}[!htbp]
    \begin{minipage}[t]{.48\linewidth}
        \centering
    
        \includegraphics[width=12.9cm,height=7.5cm,keepaspectratio]{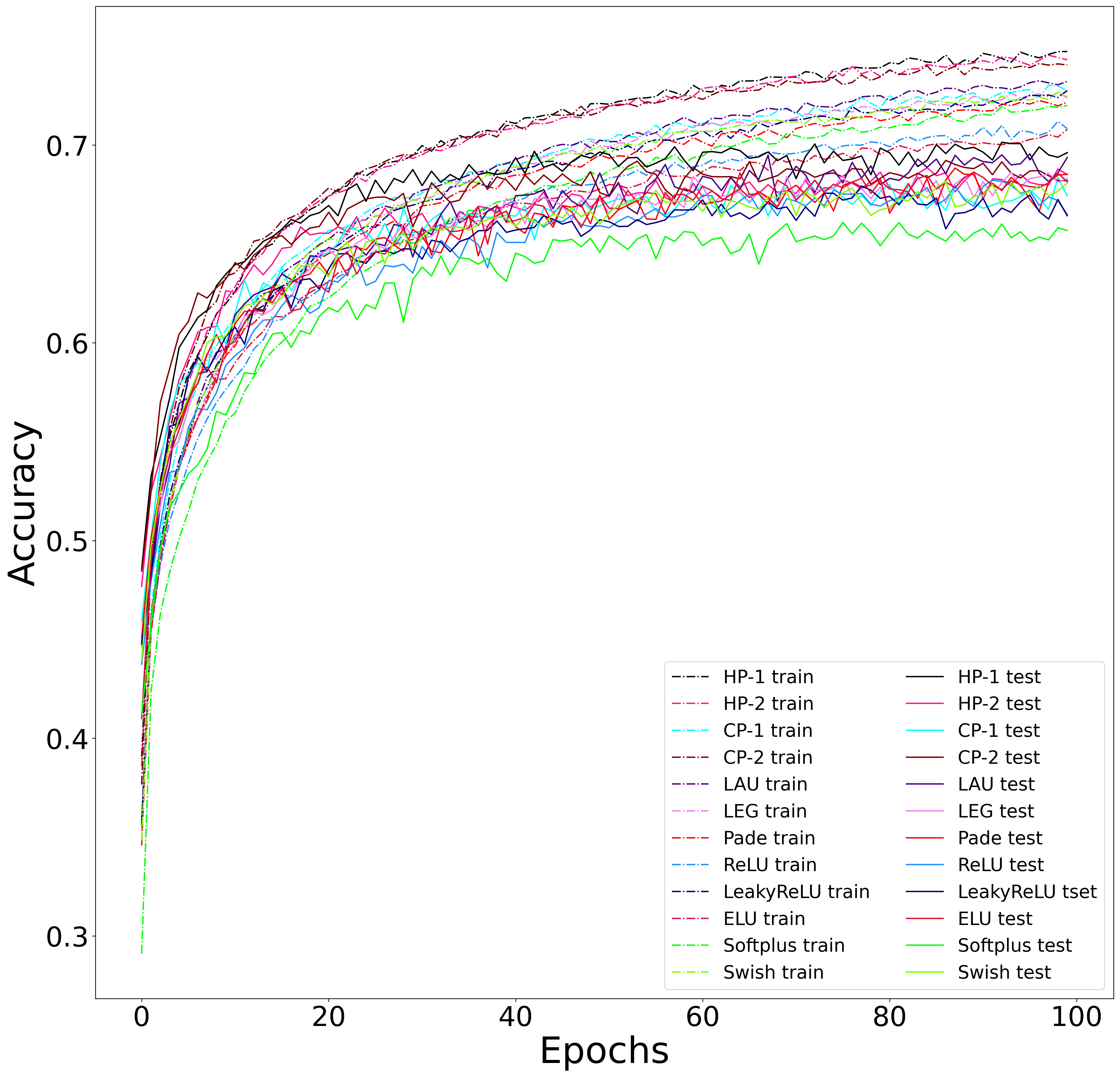}
        \caption{Top-1 Train and Test accuracy (higher is better) on CIFAR10 dataset with LeNet model for different activations}
        \label{acc1}
    \end{minipage}
    \hfill
    \begin{minipage}[t]{.48\linewidth}
        \centering
        
       \includegraphics[width=12.9cm,height=7.5cm,keepaspectratio]{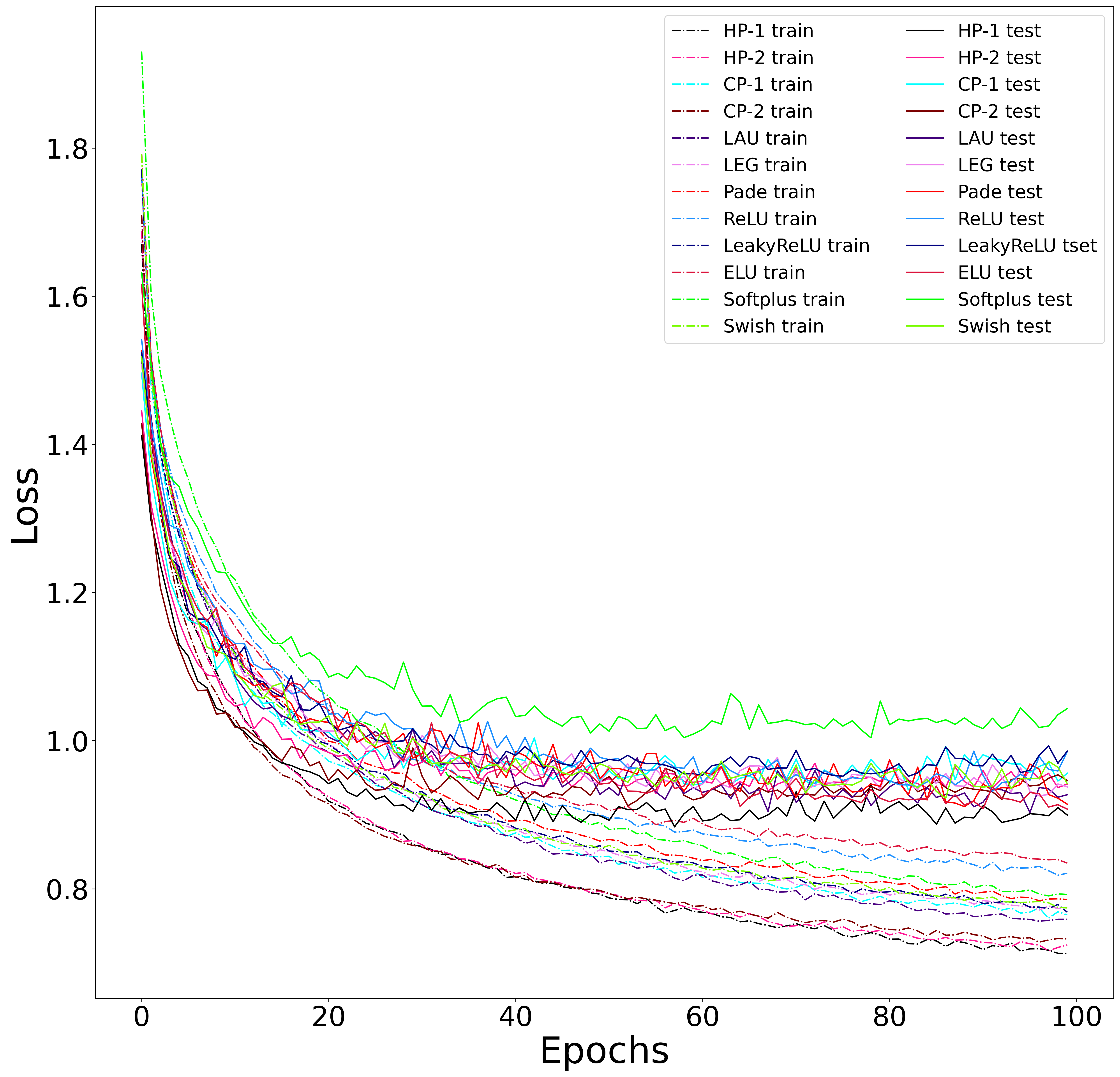}
        \caption{Top-1 Train and Test loss (lower is better) on CIFAR10 dataset with LeNet model for for different activations}
        \label{loss1}
    \end{minipage}  
\end{figure}

\begin{figure}[H]
    \begin{minipage}[t]{.48\linewidth}
        \centering
    
        \includegraphics[width=12.9cm,height=7.5cm,keepaspectratio]{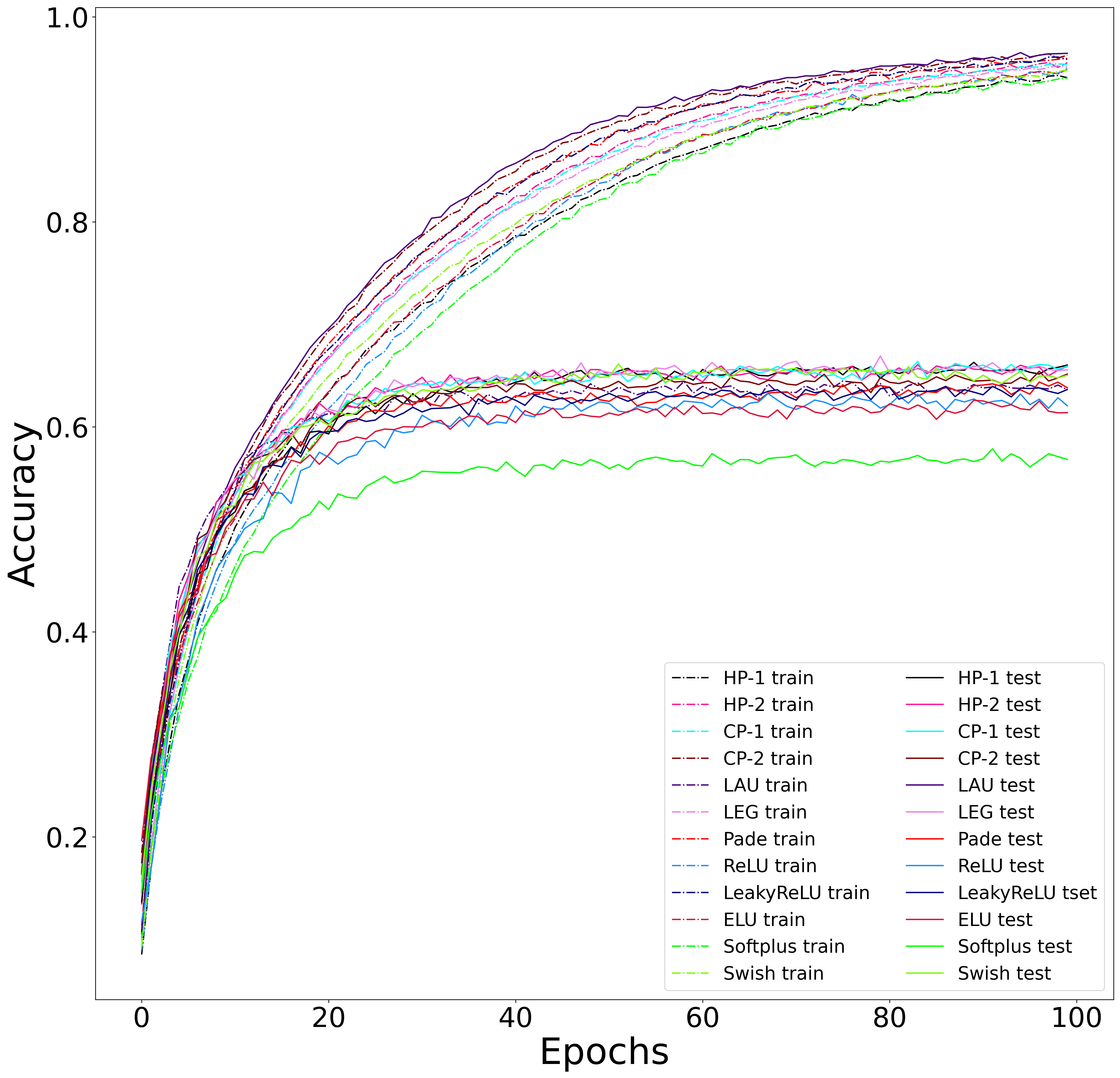}
        \caption{Top-1 Train and Test accuracy (higher is better) on CIFAR100 dataset with MobileNet V2 model for different activations}
        \label{acc2}
    \end{minipage}
    \hfill
    \begin{minipage}[t]{.48\linewidth}
        \centering
        
       \includegraphics[width=12.9cm,height=7.5cm,keepaspectratio]{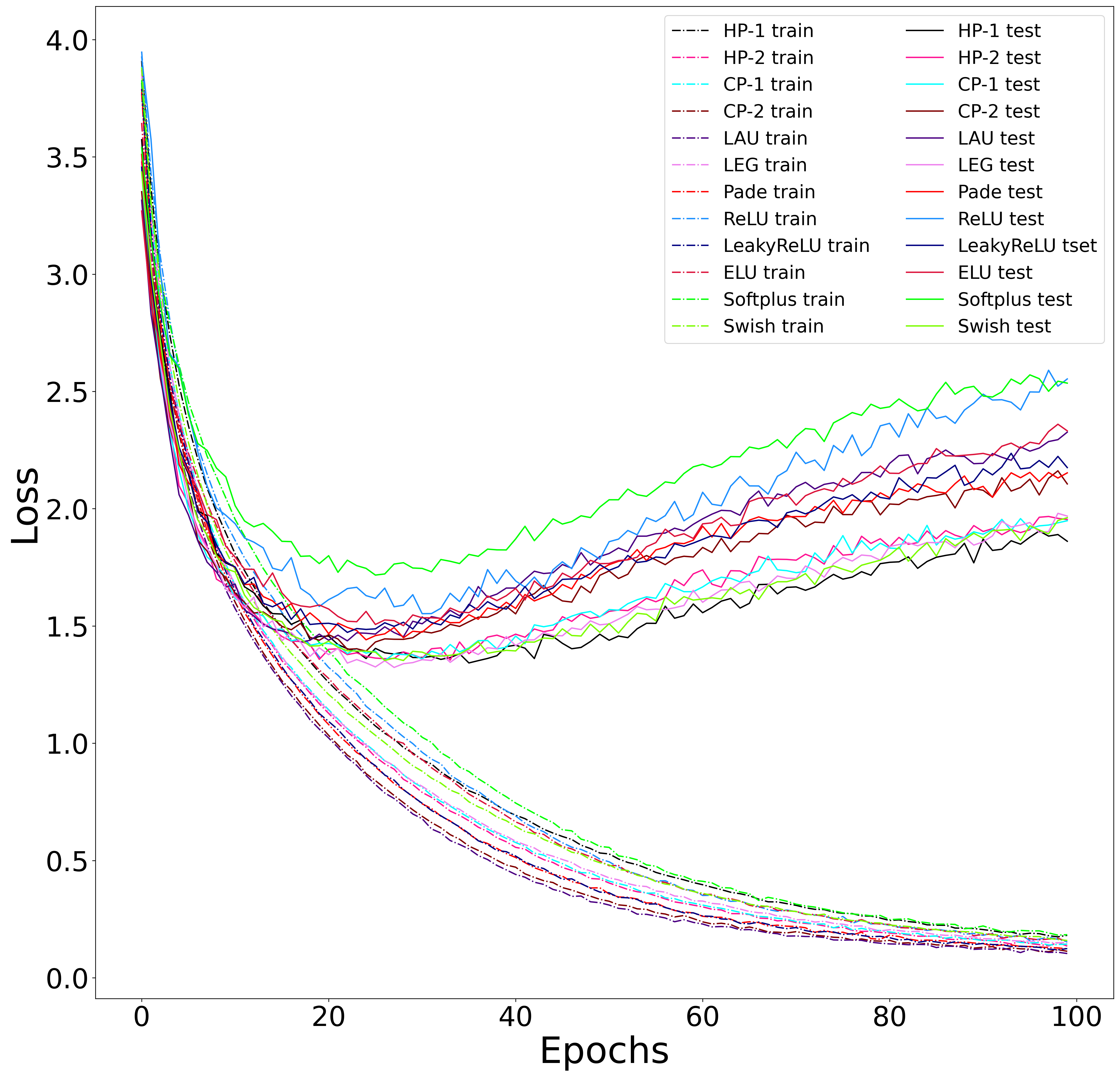}
        \caption{Top-1 Train and Test loss (lower is better) on CIFAR100 dataset with MobileNet V2 model for different activations}
        \label{loss2}
    \end{minipage}  
\end{figure}

The CIFAR \cite{cifar10} database consists of total 60k images $32\times 32$ RGB images, which is divided into 50k training and 10k test images. CIFAR dataset is divided into two database- CIFAR10 and CIFAR100. CIFAR10 database contains total 10 classes with 6000 images per class while CIFAR100 database contains total 10 classes with 600 images per class. Table~\ref{tab5} contains Top-1 accuracy for mean of 10 runs on CIFAR10 dataset and Table~\ref{tab6} contains Top-1 accuracy for mean of 10 runs on CIFAR100 dataset. results have been reported in both the database on ResNet-50 \cite{resnet}, PreActResNet-34 (PA-ResNet-34) \cite{preactresnet},  Densenet-121 (DN-121) \cite{densenet}, MobileNet V2 (MN) \cite{mobile}, Shufflenet V2 \cite{shufflenet}, Deep Layer Aggregation (DLA) \cite{dla}, EfficientNet B0 (EN-B0) \cite{efficientnet}, and Le-Net \cite{lenet} models. It evident from Table~\ref{tab5} and Table~\ref{tab6} that in most cases HP-1, and HP-2 constantly outperforms ReLU and Swish. Also, notice that there is an improvement in Top-1 accuracy from 1\% to 6\% when compared with ReLU activation in the above mentioned networks. We have considered batch size of 128, Adam optimizer \cite{adam} with 0.001 learning rate and trained the networks up-to 100 epochs. Data augmentation is used for both the datasets. Training and validation curves for ReLU, Leaky ReLU, ELU, Softplus, Swish, CP-1, CP-2, LAU, LEG, HP-1, and HP-2 activations are given in Figures~\ref{acc1}, \ref{loss1}, \ref{acc2} and \ref{loss2} in CIFAR10 and CIFAR100 dataset on Le-net and MobileNet V2 models respectively. Analysing these learning curves, it is clear that after training few epochs HP-1, and HP-2 have faster convergence capability, much stable learning, higher accuracy and lower loss when compared to ReLU. 
\begin{table}[H]
\begin{center}
\begin{tabular}{ |c|c|c|c|c|c|c|c|c| }
 \hline
\makecell{Activation\\ Function} &  MN V2 & ResNet-50 & PA-ResNet-34 & SF V2 & LeNet & DN-121 & EN-B0 & DLA \\
\hline
 ReLU  & 89.71 & 90.52 & 90.18 & 88.47 & 67.35 & 91.89 & 85.52 & 89.59\\ 
 \hline
 \makecell{Leaky ReLU \\($\alpha$ = 0.01)}  & 89.92 & 90.54 & 90.32 & 88.59 & 66.99 & 92.15 & 85.42 & 89.52\\
 \hline
 ELU  & 89.62 & 90.42 & 90.20 & 88.69 & 67.02 & 92.07 & 85.65& 89.67\\
 \hline
 Softplus  & 89.52 & 90.45 & 90.01 & 88.52 & 66.71 & 91.71 & 85.07& 89.37\\
 \hline
 Swish  & 90.07 & \textbf{90.91} & 90.98 & 89.22 & 67.95 & 91.87 & 86.27& 90.52\\
 \hline
 PAU  & 90.27 & 90.65 & 90.49 & 89.37 & 68.42 & 92.14 & 86.35& 90.59\\
 \hline
 CP-1  & 90.35 & 90.48 & 91.28 & 89.67 & 68.30 & 91.36 & 86.66& 90.49\\
 \hline
 CP-2  & 90.29 & 90.57 & 91.38 & 89.55 & 68.07 & 91.51 & 86.48& 90.57\\
 \hline
 LAU  & 90.22 & 90.53 & 91.21 & 89.51 & 68.31 & 91.81 & 86.41& 90.64\\
 \hline
 LEG  & 90.35 & 90.61 & 91.02 & 89.48 & 68.40 & 91.89 & 86.23& 90.71\\
 \hline
 HP-1  & \textbf{90.92} & 90.77 & \textbf{92.20} & \textbf{90.12} & \textbf{69.59} & \textbf{92.97} & \textbf{87.67} & \textbf{91.12}\\
 \hline
 HP-2  & \textbf{90.67} & 90.69 & \textbf{91.96} & \textbf{89.92} & \textbf{69.41} & \textbf{92.77} & \textbf{87.55} & \textbf{91.01}\\
 \hline

 \end{tabular}
 \vspace{0.2cm}
\caption{Experimental results on CIFAR10 dataset. Comparison between different baseline activations and safe OPAU activations on the basis of Top-1 accuracy(in $\%$). Mean of 10 different runs has been reported.} 
\label{tab5}
\end{center}
\end{table}
\begin{table}[H]
\begin{center}
\begin{tabular}{ |c|c|c|c|c|c|c|c|c| }
 \hline
\makecell{Activation\\ Function} & MN V2 & ResNet-50 & PA-ResNet-34 & SF V2 & LeNet & DN-121 & EN-B0& DLA \\
\hline
 ReLU &  63.20 & 64.15 & 60.39 & 61.30 & 32.62 & 67.50 & 53.02 & 60.90 \\ 
 \hline
 \makecell{Leaky ReLU \\($\alpha$ = 0.01)} & 63.54 & 64.32 & 60.51 & 61.55 & 32.94 & 67.61 & 53.25 & 61.01\\
 \hline
 ELU &  63.47 & 64.25 & 60.89 & 61.85 & 33.89 & 67.32 & 53.34 & 61.29\\
 \hline
 Softplus &  63.28 & 64.02 & 60.32 & 61.22 & 32.84 & 67.42 & 53.17 & 60.50\\
 \hline
 Swish &  63.91 & \textbf{64.87} & 63.25 & 62.26 & 34.12 & 68.07 & 54.54 & 61.70\\
 \hline
 PAU &  64.97 & 64.09 & 62.18 & 62.14 & 33.94 & 67.96 & 53.81 & 61.67\\
 \hline
 CP-1 &  65.45 & 64.41 & 64.34 & 62.34 & 34.57 & 68.32 & 54.35 & 61.74 \\
 \hline
 CP-2 &  65.32 & 64.32 & 64.49 & 62.17 & 34.18 & 68.17 & 54.28 & 61.59 \\
 \hline
 LAU & 65.64 & 64.37 & 64.01 & 62.12 & 34.01 & 68.09 & 54.21 & 61.65\\
 \hline
 LEG & 65.51 & 64.52 & 64.65 & 62.29 & 34.21 & 67.89 & 54.54 & 61.87\\
 \hline
 HP-1 &  \textbf{66.22} & 64.71 & \textbf{65.45} & \textbf{63.04} & \textbf{35.36} & \textbf{68.72} & \textbf{54.99} & \textbf{62.39}\\
 \hline
 HP-2 & \textbf{65.95} & 64.59 & \textbf{65.02} & 63.02 & \textbf{34.85} & \textbf{68.66} & \textbf{54.74} & \textbf{62.48}\\
 \hline

 \end{tabular}
 \vspace{0.2cm}
\caption{Experimental results on CIFAR100 dataset. Comparison between different baseline activations and safe OPAU activations on the basis of Top-1 accuracy(in $\%$). Mean of 10 different runs has been reported.} 
\label{tab6}
\end{center}
\end{table}
\subsubsection{\textbf{Tiny Imagenet}}
The ImageNet Large Scale Visual Recognition Challenge(ILSVRC) is considered to be one of the most popular benchmarks for image classification problems. A similar type of image classification database like ILSVRC is Tiny Imagenet, which is a smaller dataset with fewer image classes. The images in this database are of size $64 \times 64$ with total 100,000 training images, 10,000 validation images, and 10,000 test images. The database has 200 image classes with 500 training images, 50 validation images, and 50 test images in each class. A mean of 5 different runs for Top-1 accuracy is reported in table~\ref{tab22222} on WideResNet 28-10 (WRN 28-10) \cite{wrn} model. We have used a batch size of 32, He Normal initializer \cite{prelu}, 0.2 dropout rate \cite{dropout}, adam optimizer \cite{adam}, initial learning rate(lr rate) 0.01, and lr rate is reduced by a factor of 10 after every 50 epochs up-to 250 epochs. The Data augmentation method is used in this database.
\begin{table}[H]
\begin{center}
\begin{tabular}{ |c|c|c| }
 \hline
 Activation Function &  \makecell{Wide ResNet \\ 28-10 Model}  \\
 \hline
 ReLU  &  60.35  \\ 
 \hline
 Swish  &  60.69\\
 \hline
 Leaky ReLU($\alpha$ = 0.01) &  60.62 \\
 \hline
 ELU  &  60.02\\
 \hline
 Softplus & 59.81\\
 \hline
 CP-1 & 60.54\\
 \hline
 CP-2 & 60.61\\
 \hline
 LAU & 60.32\\
 \hline
 LEG & 61.31\\
 \hline
 HP-1 & \textbf{62.52}\\
 \hline
 HP-2 & \textbf{62.21}\\
 \hline
 \end{tabular}
 \vspace{0.2cm}
\caption{Experimental results on Tiny ImageNet dataset. Comparison between different baseline activations and safe OPAU activations on the basis of Top-1 accuracy(in $\%$) for mean of 5 different runs have been reported.} 
\label{tab22222}
\end{center}
\end{table}
 

\subsection{\textbf{Object Detection}}
Object Detection is considered one of the most important problems in computer vision. The Pascal VOC dataset \cite{pascal} is used for our object detection experiments. We have reported results on Single Shot MultiBox Detector(SSD) 300 model \cite{ssd} with VGG-16(with batch-normalization) network as the backbone network. We have not used any pre-trained weight in the network. The network is trained on Pascal VOC 07+12 training data and tested model performance on Pascal VOC 2007 test data. We have considered a batch size of 8, 0.001 learning rate, SGD optimizer \cite{sgd1, sgd2} with 0.9 momentum, 5$e^{-4}$ weight decay for 120000 iterations.  Table~\ref{tabod} contains a mean of 5 different runs for the mean average precision(mAP).
\begin{table}[H]
\begin{center}
\begin{tabular}{ |c|c|c| }
 \hline
 Activation Function &  \makecell{mAP}  \\
 \hline
 ReLU  &  77.2  \\ 
 \hline
 Swish  &  77.3\\
 \hline
 Leaky ReLU($\alpha$ = 0.01) &  77.2 \\
 \hline
 ELU  &  75.1\\
 \hline
 Softplus & 74.2\\
 \hline
 CP-1 & 77.3\\
 \hline
 CP-2 & 77.3\\
 \hline
 LAU & 77.2\\
 \hline
 LEG & 77.4\\
 \hline
 HP-1 & \textbf{78.0}\\
 \hline
 HP-2 & \textbf{77.9}\\
 \hline
 \end{tabular}
 \vspace{0.2cm}
\caption{Comparison between different baseline activations and safe OPAU activations on Object Detection Problem. Results are reported on SSD 300 model in Pascal-VOC dataset .} 
\label{tabod}
\end{center}
\end{table}
\subsection{\textbf{Machine Translation}}
Machine Translation is a deep learning technique in which one language is translated to another language. For our experiments, WMT 2014 English$\rightarrow$German dataset is used. The database contains 4.5 million training sentences. We have evaluated model performance on the newstest2014 dataset using the BLEU score metric. An Attention-based 8-head transformer model \cite{attn} is used with Adam optimizer \cite{adam}, 0.1 dropout rate \cite{dropout}, and trained up to 100000 steps. Other hyper-parameters are tried to keep similar as mentioned in the original paper \cite{attn}.  Mean of 5 runs has been reported on Table~\ref{tabmt} on the test dataset(newstest2014).
\begin{table}[H]
\begin{center}
\begin{tabular}{ |c|c|c| }
 \hline
 Activation Function & \makecell{
BLEU Score on\\ the newstest2014 dataset }  \\
 \hline
 ReLU  &  26.2  \\ 
 \hline
 Swish  &  26.4\\
 \hline
 Leaky ReLU($\alpha$ = 0.01) &  26.3 \\
 \hline
 ELU  &  25.1\\
 \hline
 Softplus & 23.6\\
 \hline
 HP-1 & \textbf{26.8}\\
 \hline
 HP-2 & \textbf{26.7}\\
 \hline
 \end{tabular}
 \vspace{0.2cm}
\caption{Comparison between different baseline activations and safe OPAU activations on Machine translation Problem. The results are reported on transformer model in WMT-2014 dataset.} 
\label{tabmt}
\end{center}
\end{table}
\subsection{Comparison With the baseline activation functions}
We observe that HP-1 and HP-2, in most cases, beat or performs equally well with baseline activation functions and under-performs marginally on rare occasions, and a detailed comparison of these activation functions on the basis of all the experiments provided in earlier sections is given in Table~\ref{tab49}.
\begin{table}[H]
\newenvironment{amazingtabular}{\begin{tabular}{*{50}{l}}}{\end{tabular}}
\centering
\begin{amazingtabular}
\midrule
Baselines & ReLU & Leaky ReLU & ELU & Softplus & Swish  & PAU\\
\midrule
HP-1 $>$ \text{Baseline} & \hspace{0.3cm}22 & \hspace{0.45cm}22 & \hspace{0.3cm}22 & \hspace{0.3cm}22 & \hspace{0.3cm}21 & \hspace{0.3cm}22\\
HP-1 $=$ \text{Baseline} & \hspace{0.3cm}0 & \hspace{0.45cm}0 & \hspace{0.3cm}0 & \hspace{0.3cm}0 & \hspace{0.3cm}0 & \hspace{0.3cm}0\\
HP-1 $<$ \text{Baseline} & \hspace{0.3cm}0 & \hspace{0.45cm}0 & \hspace{0.3cm}0 & \hspace{0.3cm}0 & \hspace{0.3cm}1 & \hspace{0.3cm}0 \\
\midrule
HP-2 $>$ \text{Baseline} & \hspace{0.3cm}22 & \hspace{0.45cm}22 & \hspace{0.3cm}22 & \hspace{0.3cm}22 & \hspace{0.3cm}21 & \hspace{0.3cm}22\\
HP-2 $=$ \text{Baseline} & \hspace{0.3cm}0 & \hspace{0.45cm}0 & \hspace{0.3cm}0 & \hspace{0.3cm}0 & \hspace{0.3cm}0 & \hspace{0.3cm}0\\
HP-2 $<$ \text{Baseline} & \hspace{0.3cm}0 & \hspace{0.45cm}0 & \hspace{0.3cm}0 & \hspace{0.3cm}0 & \hspace{0.3cm}1 & \hspace{0.3cm}0 \\
\midrule
\end{amazingtabular}
\vspace{0.2cm}
  \caption{Baseline table for HP-1 and HP-2. These numbers represent the total number of networks in which HP-1 and HP-2 outperforms, equal or underperforms when we compare with the baseline activation functions}
  \label{tab49}
\end{table}

\section{Conclusion}
In this paper, we propose two trainable activation functions HP-1 and HP-2, based on rational function approximation. HP-1 and HP-2 provide better experimental results on image classification and Machine translation compared to Swish, ReLU, and Leaky ReLU.

\bibliographystyle{unsrt}  
\bibliography{references}  
\newpage
\appendix
\section{Appendix}
In the following Table, We have reported the coefficients for different orthogonal polynomials reported in Table~\ref{tab1}. The coefficients are found with orthogonal polynomial basis for rational function approximation (using equation (6)) to Leaky ReLU ($\alpha = 0.01$) activation function. We have computed the orthogonal polynomials (from Table\ref{tab1} using recurrence relations) for $k=5$ and $l=4$ in equation (6). The least-square method has been adopted to optimize the error between Leaky ReLU and the rational function with orthogonal polynomials. 

\begin{table}[H]
\begin{center}
\begin{tabular}{ |c|c|c|c|c|c|c|}
 \hline
 PC & CP-1  & CP-2 & LAU & LEG & HP-1 & HP-2 \\
 \hline 
 $c_0$  & \makecell{0.4346338\\199528298} & \makecell{0.2664672\\913492625} & \makecell{1.8360445\\235354788} & \makecell{0.32073373\\302075475} & \makecell{1.1371963\\424021352} & \makecell{0.4620915\\54274137}\\ 
 \hline
$c_1$  & \makecell{0.7582218\\699682254} & \makecell{0.34803047\\019467215} & \makecell{-2.9554505\\909267266} & \makecell{0.7142799\\668606886} & \makecell{1.7979419\\128449188} & \makecell{0.4839321\\106420414} \\
 \hline
$c_2$  & \makecell{0.3178149\\433090529} & \makecell{0.1618067\\40860617} & \makecell{1.6387368\\01888696} & \makecell{0.4246816\\357328257} & \makecell{1.1020770\\550187182} & \makecell{0.1816410\\862837883}\\
 \hline
 $c_3$  & \makecell{0.05703797\\4292444685} & \makecell{0.03019799\\2889731528} & \makecell{-0.31774975\\883776296} & \makecell{0.02343409\\3682345926} &  \makecell{0.3294885\\720434351} & \makecell{0.0303762\\525152446}\\
 \hline
 $c_4$   & \makecell{0.004000911\\6269871334} & \makecell{0.00216317\\6409556791} & \makecell{-0.0239828\\18970702} & \makecell{0.00761874\\5990466922} & \makecell{0.04271857\\995060412} & \makecell{0.00207469\\0747081737} \\
 \hline
 $c_5$  & \makecell{9.93204214\\5345177e-05} & \makecell{5.442521989\\0802244e-05} & \makecell{0.01114234\\4922587972} & \makecell{0.000212053\\5423305138} & \makecell{0.002084035\\6797464945} & \makecell{5.14576205\\1699321e-05}\\
 \hline
 $d_1$ & \makecell{-0.42263720\\399740756} & \makecell{0.16740399\\142900575} & \makecell{-0.5890262\\199320808} & \makecell{0.35334130\\018360843} & \makecell{1.0846459\\888019664} & \makecell{0.24024359\\431260522}\\
 \hline
 $d_2$ & \makecell{0.1446324\\151547079} & \makecell{0.08512431\\596790718} & \makecell{-0.09392233\\765424439} & \makecell{0.21467682\\957840964} & \makecell{0.30850156\\552330404} & \makecell{0.07515668\\172628485}\\
 \hline
 $d_3$ & \makecell{-0.006010646\\6615319236} & \makecell{0.00264612\\14606926624} & \makecell{0.00391513\\9808859812} & \makecell{0.00861132\\8149930994} & \makecell{-0.04163592\\4695219075} & \makecell{0.00312816\\654786619}\\
 \hline
 $d_4$ & \makecell{0.000244052\\0667994119} & \makecell{0.000148137\\50145571406} & \makecell{0.00642035\\2790087902} & \makecell{0.000507209\\5551410509} & \makecell{0.00224051\\5203527783} & \makecell{0.000127093\\53203643316}\\
 \hline
 \end{tabular}
 \vspace{0.2cm}
\caption{Coefficient Table for Leaky ReLU rational function approximation with orthogonal basis (using equation (6)) for network initialization. 'PC' stands for Polynomial Coefficients.} 
\label{tab10}
\end{center}
\end{table}
\end{document}